\documentclass[letterpaper, 10 pt, journal, twoside]{IEEEtran}

\usepackage{cite}
\usepackage{amsmath,amssymb,amsfonts}
\usepackage{algorithmic}
\usepackage{graphicx}
\usepackage{textcomp}
\usepackage{xcolor}

\usepackage{algorithm}
\usepackage{algorithmic}
\usepackage{float}
\usepackage{bbm}
\usepackage{graphicx}
\usepackage{subcaption}
\usepackage{xcolor}
\usepackage{hyperref}       
\usepackage{url}            
\usepackage{booktabs}       
\usepackage{amsfonts}       
\usepackage{amsmath,amssymb}
\usepackage{nicefrac}       
\let\labelindent\relax
\usepackage{enumitem}
\usepackage{wrapfig}
\usepackage{multirow}

\graphicspath{{Images/}} 

\begin{document}

\title{RAMP: A Benchmark for Evaluating Robotic Assembly Manipulation and Planning}

\author{Jack Collins$^{\ast 1}$, Mark Robson$^{\ast 2, 3}$, Jun Yamada$^{\ast 1}$, Mohan Sridharan$^{3}$, Karol Janik$^{2}$, Ingmar Posner$^{1}$ 

\thanks{Manuscript received: May, 16, 2023; Revised August, 25, 2023; Accepted October, 18, 2023.}
\thanks{This paper was recommended for publication by Editor Chao-Bo Yan upon evaluation of the Associate Editor and Reviewers' comments.}
\thanks{$^{\ast}$Equal contribution.}
\thanks{$^{1}$Applied AI Lab, Oxford Robotics Institute, University of Oxford}%
\thanks{$^{2}$The Manufacturing Technology Centre, Coventry, United Kingdom}%
\thanks{$^{3}$Birmingham University, Birmingham, United Kingdom}%
\thanks{Correspondence to: {\tt\small jcollins@robots.ox.ac.uk}}%
\thanks{Project page: \url{https://sites.google.com/oxfordrobotics.institute/ramp}}%
\thanks{This work was supported by a UKRI/EPSRC Programme Grant [EP/V000748/1].}

\thanks{Digital Object Identifier (DOI): see top of this page.}
}

\markboth{IEEE Robotics and Automation Letters. Preprint Version. Accepted October, 2023}
{Collins \MakeLowercase{\textit{et al.}}: RAMP: A Benchmark for Evaluating Robotic Assembly Manipulation and Planning} 

\maketitle

\begin{abstract}

We introduce RAMP, an open-source robotics benchmark inspired by real-world industrial assembly tasks. RAMP consists of beams that a robot must assemble into specified goal configurations using pegs as fasteners. As such, it assesses planning and execution capabilities, and poses challenges in perception, reasoning, manipulation, diagnostics, fault recovery, and goal parsing. RAMP has been designed to be accessible and extensible. Parts are either 3D printed or otherwise constructed from materials that are readily obtainable. The design of parts and detailed instructions are publicly available. In order to broaden community engagement, RAMP incorporates fixtures such as April Tags which enable researchers to focus on individual sub-tasks of the assembly challenge if desired. We provide a full digital twin as well as rudimentary baselines to enable rapid progress. Our vision is for RAMP to form the substrate for a community-driven endeavour that evolves as capability matures.

\end{abstract}


\IEEEpeerreviewmaketitle

\section{Introduction}
\label{sec:introduction}
\begin{figure}
    \centering
    \vspace{6px}
    \includegraphics[width=\linewidth]{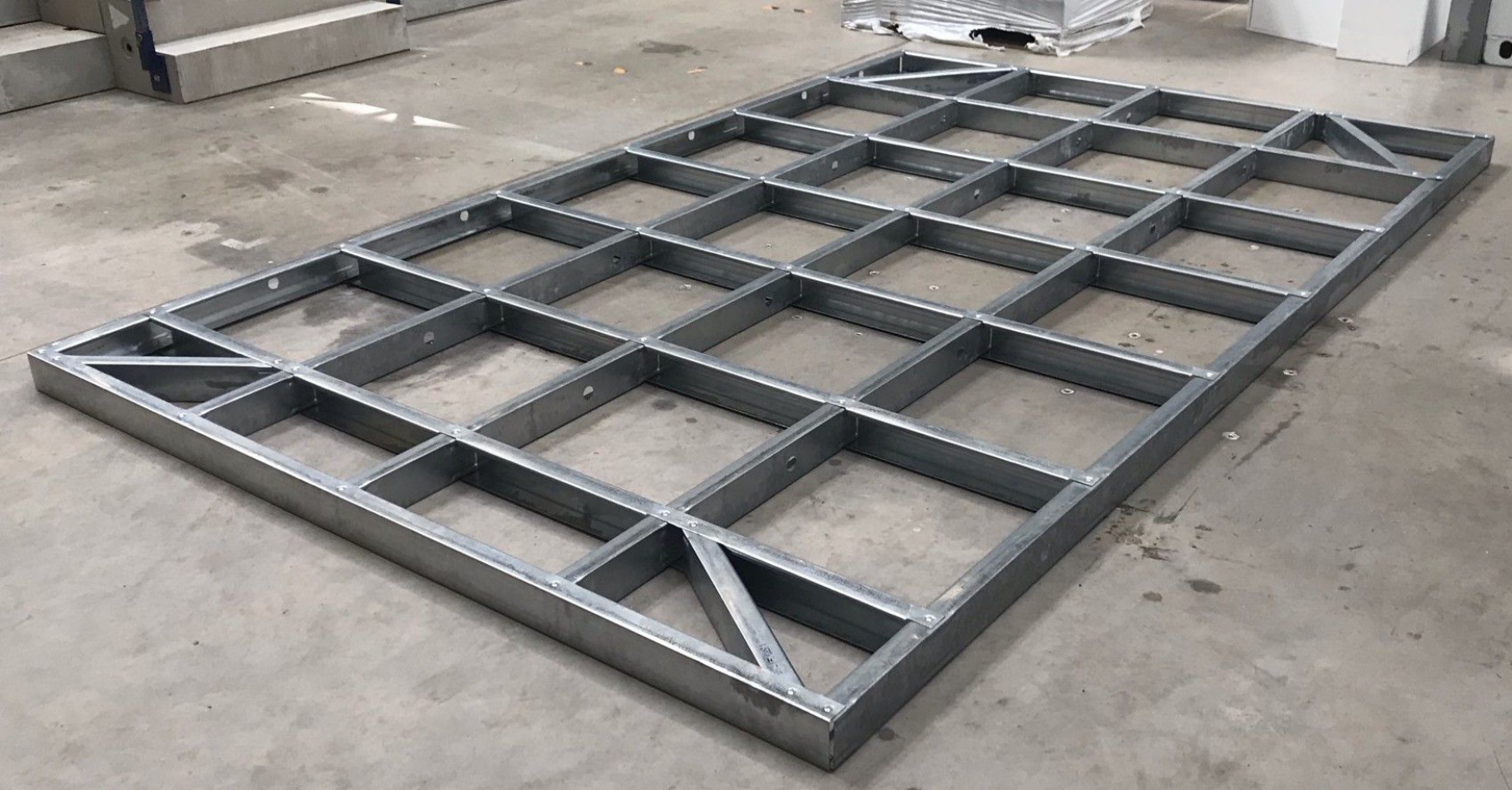}
    \caption{Light steel-rolled beams automatically cut and punched to size, and manually assembled and fastened. Such assemblies take a lot of time and effort as each beam and assembly can be unique within a prefabricated building.}
    \label{fig:industry}
\end{figure}

\IEEEPARstart{R}{obots} are predicted to raise manufacturing productivity globally by $0.8-1.4\%$ annually~\cite{manyika2017future}. Of particular interest in this domain are offsite construction tasks, where parts are largely pre-assembled offsite and only installed in-situ. Offsite construction is attractive because it supports a  substantial reduction in total build time through parallel production of modular assemblies alongside site works~\cite{smith2016off}. In the United States alone, offsite construction is predicted to increase from $33.64\%$ to $54.9\%$ in the next decade with automation levels set to increase by $7\%$ over the same period~\cite{OffsiteConstruction}. However, while beams for the internal structure of a building constructed offsite are often produced through light-steel-roll-forming before they are cut and punched to size -- a process that is readily automated -- the assembly of beams into frames remains a manual process. What complicates matters is that there is a high likelihood of every assembly being different (see Fig.~\ref{fig:industry}). Automating the assembly process itself requires robust perception, versatile skill primitives, long-horizon task and motion planning, robust action execution, and fault recovery. It touches on the full gamut of robotics research. Delivering this capability requires a focused and concerted effort. 

\begin{figure}
    \centering
    \includegraphics[width=\linewidth]{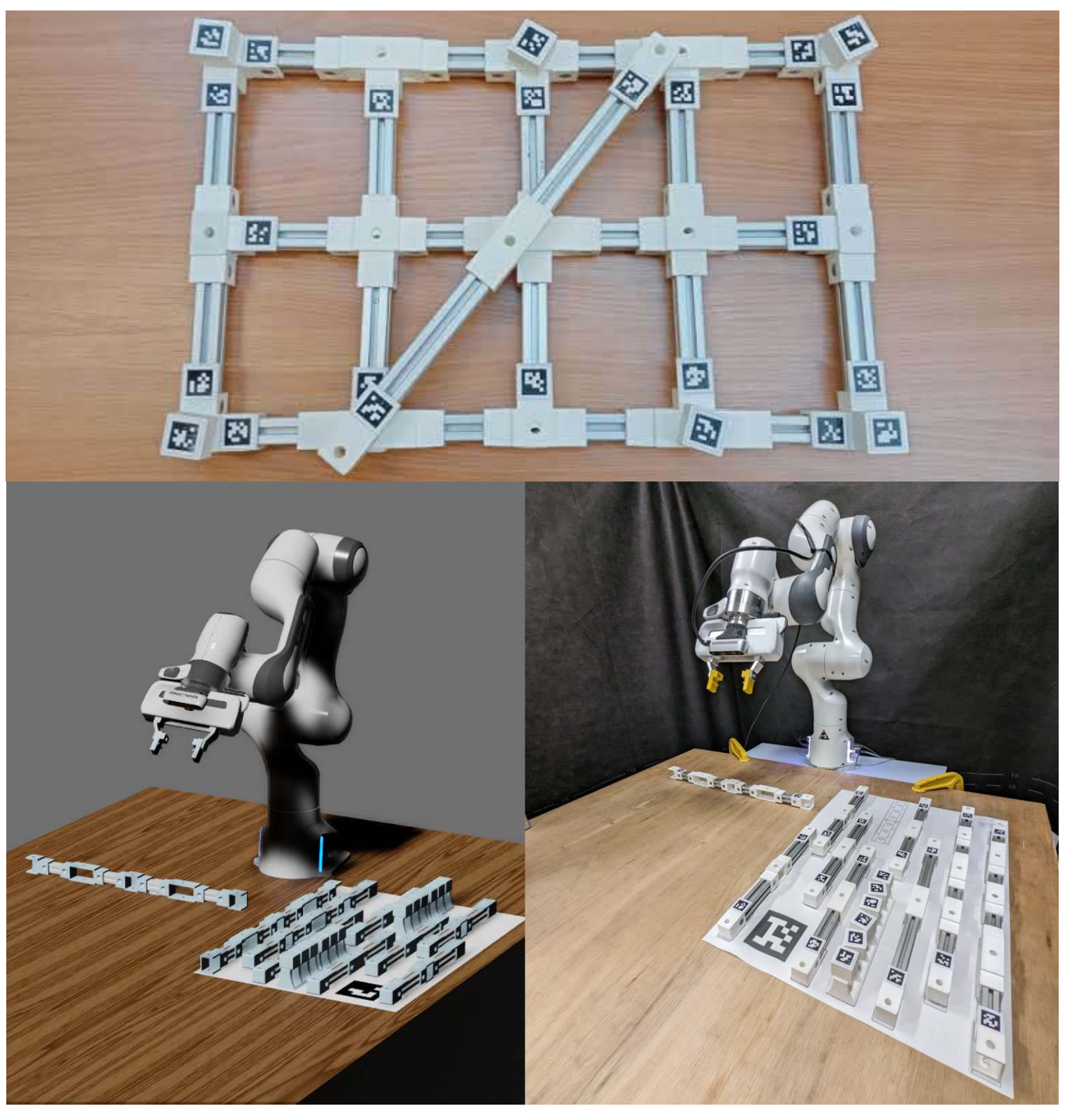}
    \caption{\emph{Top}: The benchmark beams and pegs with fiducial markers attached.  \emph{Bottom-Left}: Simulation environment with fixed beam on the left, additional beams and pegs laying on a layout template, Panda robot with optimised fingertips and wrist camera, and birds-eye camera looking down from above. \emph{Bottom-Right}: Physical setup of the benchmark with the same setup as the simulation environment.}
    \label{fig:teaser}
\end{figure}

One mechanism for fostering such focus are benchmarks. Benchmarks have had significant impact in catalysing progress across a number of different domains such as object detection~\cite{Krizhevsky2012ImagenetNetworks}, visual odometry~\cite{Geiger2012} and reinforcement learning~\cite{OpenAIGym}. However, the design of benchmarks for robotics is notoriously challenging. For one, robots are often heterogeneous across research labs, comprising many, often bespoke, interacting components and subsystems. This renders a typical like-for-like comparison of methods difficult. In addition, there is considerable disparity in the resources available across research labs working on robotics problems. Finally, real-world problems of commercial interest, such as robotic assembly, require a full systems solution, incorporating perception, planning, and control. In contrast, researchers may want to engage only in particular sub-tasks aligned with their specific research interests to drive progress. All of these factors present significant barriers to entry.

This paper presents the Robotic Assembly Manipulation and Planning (RAMP) benchmark. Situated in the space of offsite construction tasks, it is designed to mitigate the common pitfalls in robotics benchmarks mentioned above. Specifically, it is designed to be challenge-driven, accessible, and open-ended.
\textbf{Challenge-Driven.} RAMP closely mimics offsite construction challenges. It presents three categories of assemblies (\emph{easy}, \emph{medium}, and \emph{hard}) along with an evaluation protocol that reflects the need for repeatable and fast assembly of parts as well as metrics that promote speed and completeness.

\textbf{Accessible.} RAMP consists of 3D printed parts and extruded aluminium profiles (see Fig.~\ref{fig:teaser}, top), materials that are widely available. The part designs and detailed instructions are open-source and publicly available at the benchmark website\footnote{\url{https://sites.google.com/oxfordrobotics.institute/ramp/create-your-own}}. In addition, RAMP distributes with an accurate simulation environment in Nvidia Isaac~\cite{IsaacDeveloper} that mirrors the real-world setup. To enable researchers to focus on individual sub-tasks if desired, RAMP components have been instrumented to simplify scene perception. Similarly, a baseline method has been implemented both in simulation and the real-world for researchers either to build upon or to use as substrate in conjunction with their own sub-task solutions.

\textbf{Open-Ended.} While RAMP provides a number of goal configurations to work with, the parts can be used to create many other configurations. Entirely new parts can be readily created to broaden the spectrum of reachable configurations. While the parts have been designed such that goal configurations are achievable using a single arm, the RAMP benchmark readily lends itself to more complex setups such as dual-arm manipulation. In the future, we envision extensions also requiring tactile sensing and deformable object manipulation. Our vision is for RAMP to grow into a community-driven effort which evolves as capability matures.

In the following sections we review previously proposed robotic manipulation benchmarks (Sec.~\ref{sec:related-work}), explore the proposed RAMP benchmark (Sec.~\ref{sec:benchmark}), review the implementation and performance of our baseline approach (Sec.~\ref{section:baseline}) and present directions the benchmark could expand into (Sec.~\ref{sec:future}).

\section{Literature Review}
\label{sec:related-work}
Benchmarks serve an important function in stimulating research progress in key areas by enabling measurement of progress and direct comparison between methods~\cite{Calli2015BenchmarkingSet}. Although many robotic manipulation benchmarks have been proposed~\cite{Mnyusiwalla2020ASystems,Garcia-Camacho2020BenchmarkingManipulation,Calli2015BenchmarkingSet, Collins2019BenchmarkingDataset, Suarez2020BenchmarksManipulation}, relatively few focus on robotic assembly~\cite{JIANG2022102366}. Among the benchmarks for robotic assembly~\cite{Calli2015BenchmarkingSet, yu2019meta, SURREAL, Lee2021IKEATasks, Yichen}, only a subset assess long-horizon task success and real-world performance~\cite{Kimble2020BenchmarkingSystems, Chatzilygeroudis2020BenchmarkObjects}. The other robotic assembly benchmarks can be viewed as assessing performance with short-horizon tasks or in simulated environments alone.

Short-horizon robotic assembly tasks, e.g., peg-in-hole problems, have received a lot of attention with current methods achieving high success rates~\cite{Spector}. However, long-horizon robotic assembly tasks present many additional challenges. Traditionally, methods for such long-horizon tasks use prior knowledge and observations to search for a sequence of manually-encoded primitive actions (``skills") that result in the desired goal~\cite{LegoBot}. Recent work has also developed data-driven methods to jointly learn actions and plans~\cite{Ghasemipour2022BlocksLearningb}. Many frameworks for automated assembly have been investigated and assessed on a range of application spaces including Lego~\cite{LegoBot}, IKEA furniture~\cite{Suarez-Ruiz2018CanChair,Knepper2013IkeaBot:System,Lee2021IKEATasks}, timber framing~\cite{Thoma2019RoboticModules}, industrial parts~\cite{CARA}, and other profiles~\cite{Zobov2023Auto-Assembly:CAD}. As these methods for automated assembly have often been assessed on unique assembly tasks that are hard to access, it is difficult to directly compare them. Our benchmark provides a common set of easily accessible parts that existing frameworks can be evaluated upon, enabling users to directly compare methods.

Among benchmarks for multi-step real-world robotic assembly, the NIST assembly boards assess completion of a set of industry-related assembly tasks, each involving one of four defined boards with fixed goal configurations. A points-based scoring system weights components added (or subtracted), with points for parts correctly seated, inserted, clipped, threaded, and routed on a board~\cite{Kimble2020BenchmarkingSystems}. The tasks cover a range of manipulation challenges including peg-in-hole problems, gear meshing, nut threading, flexible parts handling and more. The simpler boards do not require reasoning about assembly sequencing as parts may be added or subtracted in any order, making the manipulation skills the focus of the challenge. Also, as the assembly boards are composed of many distinct parts, it can be difficult to create a new board from scratch, limiting the extent to which the benchmark is accessible and open-ended.

Another multi-step real-world robotic assembly benchmark is the bimanual semi-deformable object benchmark~\cite{Chatzilygeroudis2020BenchmarkObjects}. This benchmark is split into two separate tasks with performance assessed based on success rate and time. The first task is inspired by watchmaking; it involves assembling 3D printable parts and requires multiple interactions to insert a single part into another. The second task is rubber-band manipulation, which also requires multiple fine-motor control interactions. However, there is no requirement for high-level task planning or reasoning as the tasks are sequential with a single set goal. Additionally, there is a hardware constraint mandating the use of a bimanual platform which limits accessibility of the benchmark to research groups with such a system. 

Many competitions run at robotics conferences in past years have assessed the participants' ability to solve long-horizon assembly tasks. This includes the FetchIt! mobile manipulation challenge~\cite{FetchitChallenge}, the Industrial Assembly Challenge run multiple times at the World Robot Summit~\cite{Yokokohji2019Assembly2018} and the Robotic Grasping and Manipulation Competition that uses the NIST assembly boards~\cite{nist_2022}. Such competitions provide a motivating, competitive setting for research groups to assess performance on a common set of tasks, oftentimes using the same hardware. However, although competitions form a good basis for a benchmark assessing long-horizon robot assembly, the parts are not easily accessible to non-competitors and therefore there remains a need for an open-source, accessible alternative. 

Common themes across the reviewed robotic benchmarks and competitions include a lack of widely accessible benchmarking components, and systems that assess the ability of a method to achieve a fixed goal at the expense of generalisation to new, similar goal configurations. Our benchmark, RAMP, seeks to overcome these limitations of existing benchmarks and competitions while building on their desirable attributes, to make three key contributions. First, RAMP promotes accessibility by developing parts that are easy to create. Second, it is designed to be open-ended by providing parts that are reconfigurable and expandable such that the same parts can be used to assemble many unique designs, i.e., it can be used to assess the ability of a solution to generalise to new goal configurations. Finally, RAMP is designed explicitly to assess long-horizon, multi-step reasoning on physical systems, a challenging problem in robotics research.

\section{Benchmark} 

\label{sec:benchmark}
RAMP is designed to jointly assess planning and execution for robotic assembly, with the core principles of RAMP to provide a challenge-driven, accessible and open-ended benchmark. Currently, the benchmark is intended to assess single-arm manipulation although we hope to see it grow with the community to encompass other domains. The benchmark consists of a set of predefined beams made from an extensible set of base parts, a predefined set of assemblies for comparable assessment, metrics for assessing performance, an evaluation protocol to promote reproducibility, and an accurate simulation environment for prototyping and testing. Each of these components is described further below. 

\subsection{Base Parts}
\label{sec:benchmark_components}
The benchmark is composed of nine base beams and up to $15$ pegs---see Fig.\ref{fig:teaser} (top). The beams are constructed from $20 \times 20$ mm extruded-profile aluminium with 3D-printed joints. To construct the base beams, $33$ 3D joints need to be printed with $24$ lengths of aluminium extrusion cut to size. To assist with state estimation and beam identification, set joints have space for a $25 \times 25mm$ April Tag~\cite{Olson2011AprilTag:System}. The joints' part files and an exact bill of parts detailing the number and type of joints to print, the lengths of aluminium profile, and an assembly guide are available on the RAMP website. 

In addition to the base beams, other parts are available that can be 3D printed including fingertips designed for picking up the beams, clamps to help fix beams to a table, and holders that position pegs at a convenient pick orientation. To assist with localisation of parts and reproducible research, an A2 template is available for laying out parts, with an outline for all beams and pegs, and a large fiducial marker for state estimation. We anticipate the number of beams and assemblies to evolve to include more challenging designs over time.

\begin{figure*}[htb!]
  \centering
  \vspace{6px}
  \begin{subfigure}{\linewidth}
    \centering
    \includegraphics[width=\linewidth]{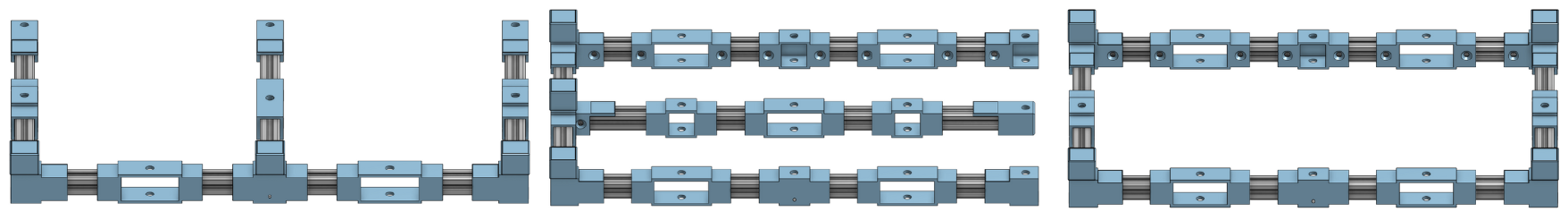}
    \caption{Easy Assemblies}
    \label{fig:subplot1}
  \end{subfigure}
  
  \begin{subfigure}{\linewidth}
    \centering
    \includegraphics[width=\linewidth]{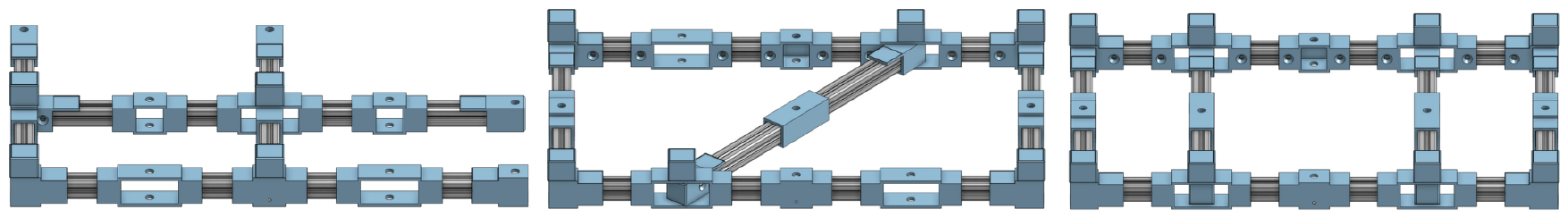}
    \caption{Medium Assemblies}
    \label{fig:subplot2}
  \end{subfigure}
  
  \begin{subfigure}{\linewidth}
    \centering
    \includegraphics[width=\linewidth]{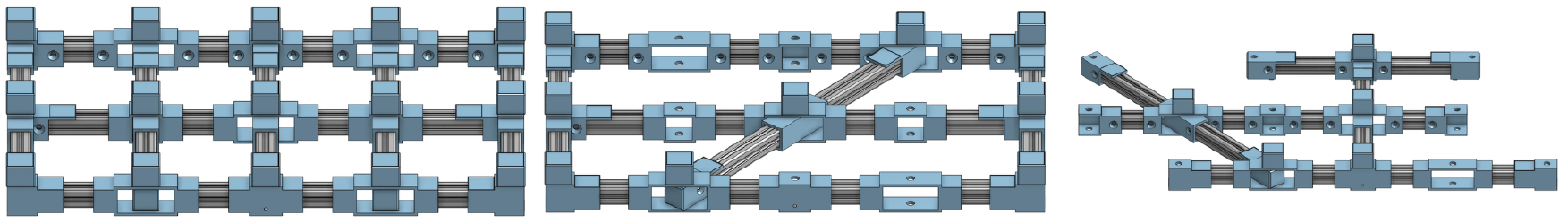}
    \caption{Hard Assemblies}
    \label{fig:subplot3}
  \end{subfigure}
  
  \caption{Three classes of assemblies for assessing the capabilities of any proposed solution to the benchmark.}
  \label{fig:goal}

\end{figure*}

\subsection{Tasks} 
\label{section:tasks}
RAMP's goal configurations are grouped into three classes: easy, medium, and hard. Fig.~\ref{fig:goal} shows the goal configurations for each class of assemblies, with more sophisticated skills being required to complete the medium and hard classes. For example, the easy class of problems only contains $3-4$ peg insertions with the most complicated action being a capping action to complete a rectangle. For the medium class of problems, $4-8$ pegs need to be inserted, with additional skills necessary to insert a beam at an angle, slide beams through other beams, and to re-grasp beams during such actions. For the hard class of problems, there are free-form designs that require additional planning considerations and beams that need to be angled suitably before sliding them through other beams. In general, the hard class of assemblies requires a longer sequence of actions which makes them difficult to complete successfully.

In addition to the image of the goal configuration, the configurations of the beams in the goal state are described using Extensible Markup Language (XML). The XML format is easy to use with existing planners, and this process can be automated; other modalities (e.g., images, natural language) are possible and encouraged.

When taken together, these three classes of assemblies pose fundamental open problems in perception, state estimation, long-horizon planning, control and execution of primitive skills, fault recovery, and learning. These goal configurations should act as a holdout set with similar designs also achievable using the same methods.

\subsection{Performance Measures}
\label{sec:metrics}
The performance of methods being tested on the the RAMP benchmark is assessed using measures that have been chosen carefully to reflect the needs of industry. The two priorities are that the assemblies: (i) are completed accurately; and (ii) in the shortest amount of time. An insertion is accurate if a peg is inserted, such that no stem is visible, into the correct hole pinning the required two joints of the beams. If a peg is not inserted into the correct hole, is not flush with the surface of the joint, or is not pinning the correct two joints together then the task completion percentage does not increase. The results obtained from RAMP thus include the time of each successful peg insertion with the time also including the time taken to generate the executed plan. These results are presented as plots similar to those found in Fig.\ref{fig:plots}, and include the average across five repeats and the most successful attempt.

\subsection{Evaluation Protocol}
\label{sec:protocol}
For reproducibility, we strongly suggest users adhere to our evaluation protocol. Specifically, an entire class of assemblies must be completed using the same code, with only the goal changing between assemblies. To assess robustness and repeatability of the methods being evaluated, each assembly in a class needs to be performed five times consecutively without altering the code; the only change between individual runs, as stated before, is the goal to be achieved. 

To promote accessibility, RAMP is designed to be used with any robot that has a manipulator arm and sensors for state estimation. The manipulator can also use any available gripper ranging from parallel grippers to anthropomorphic hands. Diverse set of sensor setups are encouraged for state estimation as this promotes the exploration of different solution methods for the underlying perception challenges. 

To support the use of a single manipulator, the first beam of the assembly is fixed rigidly to the table, with all assemblies in the benchmark using it as the starting point. The fixed beam must be located within the robot's operational space. All other parts for any target assembly must start on the A2 template in their allocated spot. The A2 template must not lie below the completed assembly but no other constraints are imposed on its starting location; Fig.~\ref{fig:teaser} (bottom right) shows an example setup. From the start state, the assembly must be constructed using the available parts. More than the necessary number of pegs may be used; if a peg is dropped during assembly, another peg may be used to achieve the goal. 

\subsection{Simulation}
\label{sec:simulation}
\begin{figure}
    \centering
    \vspace{6px}
    \includegraphics[width=\linewidth]{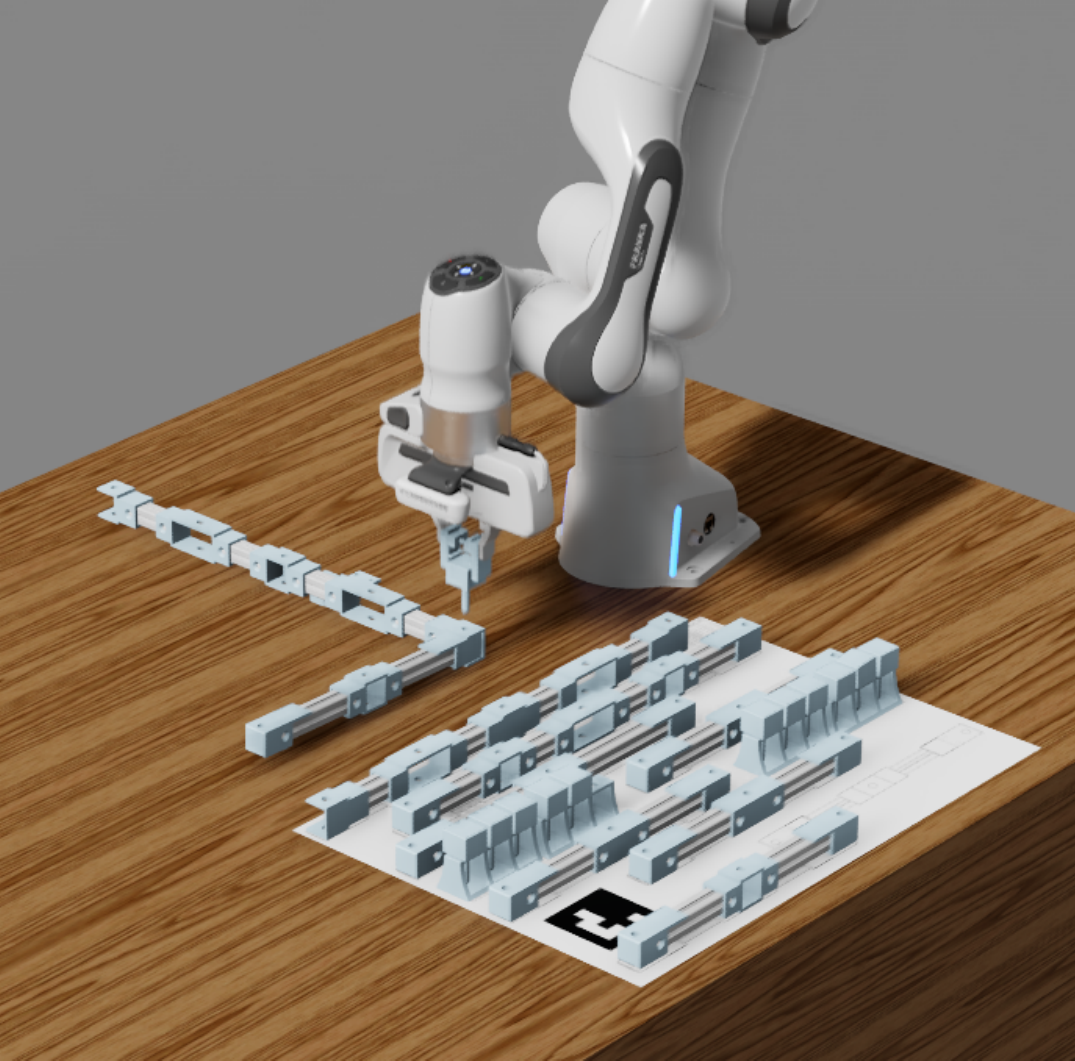}
    \caption{The open-source RAMP simulation environment using Nvidia Isaac. Assembly in simulation uses SDF contacts.} 
    \label{fig:simulation}
\end{figure}

RAMP includes a complete simulation environment created to mirror the real-world setup (see Fig.~\ref{fig:simulation}). The environment is built using the Nvidia Isaac simulator~\cite{IsaacDeveloper} which comes with an API, real-time physics simulation capability, and realistic renderings. Faithful collision detection based on Signed Distance Fields (SDF)~\cite{Macklin2020LocalCollision} is used to compute fine-resolution contacts between the gripper (of the manipulator) and the beams, and for reliably simulating the insertion of joints and pegs.

The simulation environment unlocks access to a broad range of learning-based methods, and enables development and assessment of solutions without needing access to a real-world setup. The simulation provides observations including end-effector pose and force-torque data; joint position, velocity and torque; peg and beam state information; and image observations for wrist and birds-eye cameras. We have also created high-level interfaces to basic control approaches including Cartesian position and impedance control. 

We anticipate multiple extensions of the RAMP simulation environment. This includes increasing the realism of the environment through the addition of observation noise and the integration of April Tags for marker-based state estimation. Additional control methods, e.g., force-based and closed-loop policies, are another opportunity to improve the environment. Finally, to expand the environment to learning-based methods, the inclusion of procedurally generated beams and assemblies as well as the ability to parallelise the environment are important directions for future work.

\section{Baseline Implementation} \label{section:baseline}
RAMP includes a baseline method that aims to lower the barrier to entry for new adopters and can serve as a foundation for new methods. The baseline pursues a classical approach to solve the easy class of assemblies with the integration of a task planner, conditioned on a goal assembly specified in XML, and hand-engineered skills to execute the derived plan. The baseline's components and performance are discussed below.

\subsection{Task Planner}
The baseline task planner is a simplification of an existing refinement-based architecture (REBA) that reasons with any given domain's transition diagrams at two resolutions, with the fine-resolution description defined formally as a \textit{refinement} of the coarse-resolution description~\cite{Sridharan2019REBA}. In REBA, for any given goal, non-monotonic logical reasoning at the coarse-resolution with commonsense domain knowledge computes a plan of abstract actions. Each abstract transition is implemented as a sequence of fine-resolution actions by automatically \textit{zooming} to and reasoning (probabilistically) with the relevant part of the fine-resolution description. 
The fine-resolution outcomes are then mapped to abstract ones that are used for subsequent reasoning. REBA has enabled reliable, efficient, and transparent decisions on robots and simulated agents in complex domains as it supports inference, planning, and diagnostics in the presence of noisy sensing and actuation. 

\begin{figure*}[ht!]
    \centering
    \includegraphics[width=0.9\linewidth]{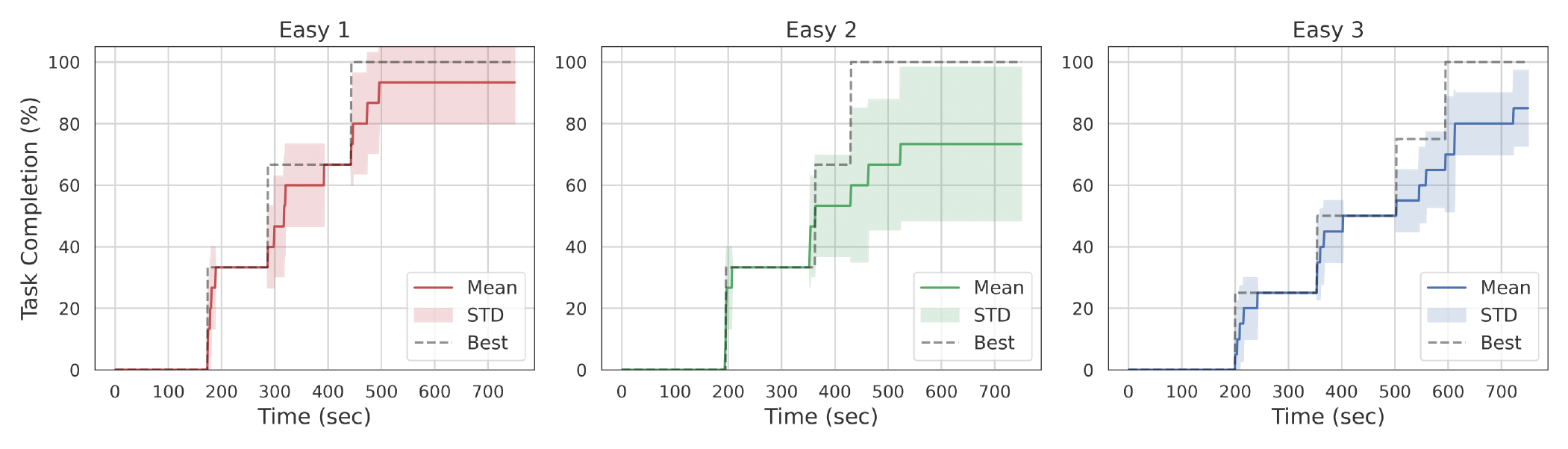}
    \caption{Results of the baseline method across three assemblies in the easy class of the benchmark, with the percentage of completion of the desired assembly expressed as a function of the time (planning and execution, in seconds). The best and average of the repeats are plotted along with the standard deviation. Standard deviation increases with execution time since the more complex assemblies typically require more actions.} 
    \label{fig:plots}
\end{figure*}

In our baseline, the coarse-resolution system description $\mathcal{D}_c$ comprises statements in action language $\mathcal{AL}_d$~\cite{Gelfond2013ALD} encoding the domain signature ($\Sigma_c$) and axioms governing domain dynamics. $\Sigma_c$ includes a hierarchy of \emph{sorts} such as $robot$, $place$, $thing$, $object$, $beam$, and $step$ (for temporal reasoning); actions such as $move(robot, place)$, $pick\_up(robot, beam)$, and $assemble(robot, beam)$; 
statics, i.e., domain attributes such as $next\_to(place, place)$ that cannot be changed; and fluents, i.e., domain attributes that can be changed, e.g., $loc(thing, place)$ and $in\_hand(robot, thing)$. Axioms in $\mathcal{D}_c$ include \emph{causal laws}, \emph{state constraints}, and \emph{executability conditions} exemplified by the following examples:
\begin{align*}
    &putdown(R, O)~\textbf{causes}~\lnot~in\_hand(R, O)\\
    &loc(O, Pl)~\mathbf{if}~loc(R, Pl),~in\_hand(R, O)\\    
    &\mathbf{impossible}~pick\_up(R, O)~\mathbf{if}~in\_hand(R, O)
\end{align*}
where variables \textit{R}, \textit{O}, and \textit{Pl} represent the robot, an object, and a place respectively. 

To compute a plan of abstract actions to achieve a given goal, our baseline constructs and solves a program $\Pi(\mathcal{D}_c, \mathcal{H}_c)$ in Answer Set Programming (ASP)~\cite{gelfond2014ASP} that encodes $\mathcal{D}_c$, the initial conditions, a history $\mathcal{H}_c$ of prior observations and action executions, and heuristic methods and helper axioms that direct the search for the plan. ASP encodes concepts such as \textit{default negation} and \textit{epistemic disjuction}, supports non-monotonic logical reasoning, and reduces reasoning to computing \textit{answer set(s)} of the corresponding program. Each answer set of $\Pi(\mathcal{D}_c, \mathcal{H}_c)$ includes a sequence of abstract actions to reach the goal. To compute answer sets, we use the SPARC system~\cite{Balai2013SPARC} and the \textit{clingo} SAT solver~\cite{Gebser2019MultishotASPSWC}. 

To map abstract transitions to actions that can be executed by the robot, our baseline supports a fine-resolution system description $\mathcal{D}_f$, a refinement of $\mathcal{D}_c$, which can be viewed as looking at $\mathcal{D}_c$ through a magnifying lens to discover structures that were previously abstracted away. The sorted signature of $\mathcal{D}_f$ includes sorts for these new structures and those in $\mathcal{D}_c$, and actions, statics, and fluents described in terms of these sorts, e.g., $assemble(robot, object\_part)$ and $in\_hand(robot, object\_part)$. The axioms are refined in a similar manner, and additional \emph{bridge axioms} link the attributes in $\mathcal{D}_c$ and $\mathcal{D}_f$. For each abstract transition, $\Pi_f(\mathcal{D}_f, \mathcal{H}_f)$ is solved to extract answer sets with sequences of fine-resolution actions. The planning code for our baseline is available through the RAMP website\footnote{\url{https://sites.google.com/oxfordrobotics.institute/ramp}}.

The ordered list of fine-resolution actions output by the planner are to be executed by the robot in an open loop. Specifically, the following assembly actions are used:
\begin{itemize}
    \item \emph{move(robot, place)}: to move the end effector to a location.
    \item \emph{pick\_up(robot, part)}: to grasp a part with the gripper.
    \item \emph{put\_down(robot, part)}: to release a part from the gripper.
    \item \emph{assemble\_square(robot, joint)}: to insert beams where a specified joint fits into an existing joint in the assembly.
    \item \emph{assemble\_cap(robot, joint)}: to insert a beam such that one or more joints in the assembly fit into the beam.
    \item \emph{fasten(robot, joint, joint, peg)}: to insert a peg into two beams at their connecting joints.   
    \item \emph{push(robot, beam)}: to correct the position of beams already in the assembly.
\end{itemize}
The implementation of these actions is discussed below.

\subsection{Action Execution}
\label{sec:action-execute}
Any given plan is executed using hand-designed skills that are intentionally designed to act as the simplest approach. Each fine-resolution action has a corresponding skill that is executed without re-planning. Sensors (e.g., wrist-mounted and bird's eye cameras) are used to observe the April tags~\cite{Olson2011AprilTag:System} on the beams and pegs, estimating their location in world coordinates. Since the offsets of these tags to each joint and link of the beams are known, joint and link poses can be computed in the global reference frame.

Our baseline uses the \emph{MoveIt}~\cite{gorner2019moveit} motion planner for executing simple actions such as \emph{move}, \emph{pick\_up}, and \emph{put\_down}, and uses Cartesian-space impedance control and force feedback for executing contact-rich actions.
For example, to complete the peg-in-hole skill for the assembly action \emph{fasten}, force feedback is essential to detect any contact between the peg and the beam surface. The robot moves downwards until a specific z-axis force threshold is reached, indicating contact between the peg and the beam surface. A spiral search algorithm~\cite{chhatpar2001search} is then employed to locate the hole position. After the peg is aligned with the hole, the robot repeatedly moves forward and backward to prevent jamming during insertion. If unsuccessful in detecting the hole, the robot moves upwards to re-estimate the hole position using the nearest fiducial marker observed by the wrist camera and re-attempts the insertion.

The \emph{assemble\_square} assembly action inserts the grasped beam into another beam already in the assembly. This skill requires the robot to move linearly until the grasped beam makes contact with the other beam, detected using force feedback. Then, the joint of the other beam in the assembly is localised by moving the grasped beam along the face of contact in a 1D implementation of the spiral search algorithm. After localising the joint in the assembly, the robot aligns the two beams using force feedback.

The assembly action of \emph{assemble\_cap} similarly uses Cartesian impedance control to move the grasped beam towards the desired pose until a reactionary force, exerted against the robot, exceeds a threshold value. After capping, alignment of the beams is achieved by the assembly action \emph{push}.
The \emph{push} action corrects the position of a beam aiming to reposition an inserted beam to the nominal target position in order to ensure alignment with future beams that might be inserted into the assembly. This skill is implemented using Cartesian impedance control and executed until a force threshold is reached.

\begin{figure*}[ht!]
    \centering
    \vspace{6px}
    \includegraphics[width=\textwidth]{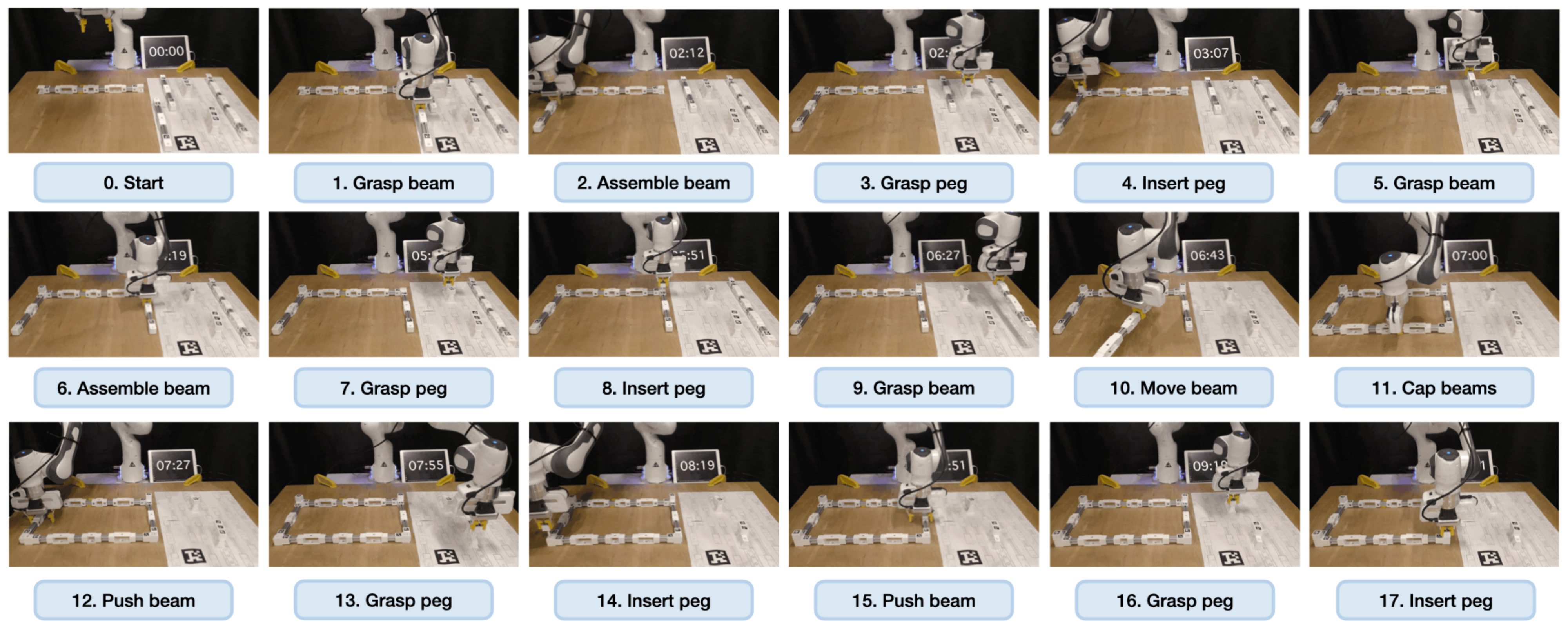}
    \caption{Visualisation of a plan attempted by a robot using the baseline method for the third easy assembly from Fig.\ref{fig:subplot1}.} 
    \label{fig:timelapse}
\end{figure*}

\subsection{Results}
\label{sec:discussion}
We evaluated the baseline method using the evaluation protocol from Section~\ref{sec:benchmark}, with the results presented in Fig.~\ref{fig:plots}. Fig.~\ref{fig:timelapse} shows snapshots of the actions and plan adopted by the baseline for an assembly, and further videos are available on the RAMP website. Fig.~\ref{fig:plots} depicts the proposed metric of task completion percentage versus time (in seconds) across the three easy assemblies. Each subplot plots the average, best and standard deviation over the five repeats. To summarise the performance of our baseline, we compute the average of the mean successes and the mean time across the three assemblies, which is $84\%/580sec$. We observe that planning takes $180-200$ seconds. Given the simplicity of the baseline, we expect more sophisticated methods to achieve a higher success rate in a shorter period of time. 

A significant portion of failure cases in these experiments stem from the peg insertion action. The robot struggles to accurately estimate the hole due to observation noise and is unable to correct for slanted pegs held within the gripper. Also, when the robot fails to achieve precise alignment of two beams, insertion of the peg into the designated hole is not possible. The peg insertion action is one of the more difficult actions within the benchmark but could be made more reliable by modeling and reasoning about the uncertainty in perception and actuation, and by introducing learning-based methods like InsertionNet that provide high success rates on challenging insertion problems~\cite{Spector}. Fig.~\ref{fig:plots} also shows that the standard deviation increases with time, implying that assemblies that require a greater number of actions result in lower success rates. Improving the integration between perception, planning and action execution is essential to improve the observed performance on the easy class of assemblies and to consistently complete the more complex assemblies in the medium and hard classes of RAMP. This can be done by improving the baseline method or by developing other sophisticated methods. Furthermore, to attempt the medium and hard classes of RAMP, the set of assembly actions used by the planner needs to be extended, e.g., to handle rotated beams, and all fine-resolution actions will need to be enhanced to achieve a higher success rate.

\section{Looking Ahead}
\label{sec:future}
There are many opportunities to improve upon the proposed baseline implementation to overcome the limitations and see vast improvements in the results. One such improvement could be to extend the task planner to support diagnostics in order to recover from failure cases by detecting faults and re-planning. Further to this point, tighter coupling between the task planner and the motion planner in the form of a better integration of  the continuous representations necessary for robot control and the discrete representations used for higher-level task planning, would lead to improvements in the generation and successful execution of the sequences of skills for various tasks~\cite{garrett:ARCRAS21}.

There are many avenues to increase the robustness and speed of the manipulation skills, such as by using learned motion plans~\cite{yamada2023leveraging}, and combining motion planning with learned contact-rich manipulations~\cite{yamada2023efficient,yamada2020mopa}. Integrating additional modalities of sensing could also see improvements in skill reliability, with one promising option being tactile sensing as touch is a crucial sensory input relied upon by humans to assemble similar objects~\cite{Lambeta_2020}.

RAMP supports easy extension to additional open challenges in robotics, some of which we discuss here. The benchmark is currently designed to be completed by a single-arm system but it could easily be extended to assess multi-agent planning and dual-arm coordination~\cite{Chatzilygeroudis2020BenchmarkObjects} by removing the rigidly attached beam and using systems with multiple manipulators. Another important extension is to assess disassembly rather than assembly as the search space for assembly can be reduced by first solving the disassembly problem~\cite{AssembleThemAll}. The assembly problem can also be extended to 3D by designing additional parts that build upon the base set of parts, this would bring with it additional challenges in path planning and collision avoidance. In addition, extending the simulation environment as discussed in Sec.~\ref{sec:simulation} would open additional opportunities for a separate virtual benchmark and enable learning-based assembly methods. Finally, removing the fiducial markers from the beams would extend the benchmark to include challenges in object identification and pose estimation whilst making the benchmark a more accurate representation of the real-world problem.

\section{Conclusion}
RAMP is a challenge-driven, accessible, and open-ended benchmark. It is industry-insipred with the objective of catalysing the research community toward making progress on open challenges related to robot assembly. To make the benchmark easily accessible, the physical components are open-sourced and made from easy-to-source 3D printed materials and extruded aluminium profiles. These same parts can be reconfigured into any number of designs, making the range of assemblies open-ended. RAMP is distributed with a rudimentary baseline implemented both on real-world hardware and in an accurate simulation environment. Adopters of RAMP can choose to develop a full systems solution or build upon the provided baseline to drive progress in areas aligned with their interests. In the future, we plan to further develop the proposed baseline method and to expand the benchmark to support other settings such as bimanual manipulation. Our vision is for RAMP to form the foundation for a community-driven endeavour, which evolves as capability matures.

\section*{Acknowledgement}

For the purpose of open access, the authors have applied a Creative Commons Attribution (CC BY) license to any Accepted Manuscript version arising.

\bibliographystyle{IEEEtran}
\bibliography{references.bib}

\end{document}